\pdfoutput=1
\documentclass[letterpaper, 10 pt, conference]{ieeeconf}  

\IEEEoverridecommandlockouts                              

\usepackage{cite}
\usepackage{graphics} 
\usepackage{graphicx}
\usepackage{amsfonts}
\usepackage[normalem]{ulem}
\usepackage[caption=false,font=footnotesize]{subfig}
\usepackage{amsmath} 
\usepackage{color}
\usepackage{hyperref}
\hypersetup{
    colorlinks=black,
    linkcolor=black,
    filecolor=black,      
    urlcolor=black,
}

\title{\LARGE \bf
Design and Control of Roller Grasper V2 for In-Hand Manipulation
}

\author{Shenli Yuan$^{1,2}$, Lin Shao$^{2}$, Connor L. Yako$^{1,2}$, Alex Gruebele$^{1}$, and J. Kenneth Salisbury$^{2}$
\thanks{*This work has been funded, in part, by the Stanford Interdisciplinary Graduate Fellowship.  Detailed implementations can be found at: \url{https://ccrma.stanford.edu/\~shenliy/roller_grasper_v2.html}}
\thanks{$^{1}$Department of Mechanical Engineering, Stanford University}%
\thanks{$^{2}$Stanford Artificial Intelligence Lab
(SAIL), Stanford University}%
\thanks{{\tt{\{shenliy, lins2, clyako, agruebe2\}@stanford.edu, jks@cs.stanford.edu}}}
}

\begin{document}

\maketitle
\thispagestyle{empty}
\pagestyle{empty}

\begin{abstract}
\label{sec:abstract}
The ability to perform in-hand manipulation
still remains an unsolved problem; having this capability would allow robots to perform sophisticated tasks requiring repositioning and reorienting of grasped objects. In this work, we present a novel non-anthropomorphic robot grasper with the ability to manipulate objects by means of active surfaces at the fingertips. Active surfaces are achieved by spherical rolling fingertips with two degrees of freedom (DoF) -- a pivoting motion for surface reorientation -- and a continuous rolling motion for moving the object. A further DoF is in the base of each finger, allowing the fingers to grasp objects over a range of size and shapes. Instantaneous kinematics was derived and objects were successfully manipulated both with a custom handcrafted control scheme as well as one learned through imitation learning, in simulation and experimentally on the hardware.

\end{abstract}

\section{Introduction}
\label{sec:intro}

In an effort to bring robots from the laboratory into real world environments, researchers have endeavored  to develop increasingly dexterous robots that can interact deftly with objects. In order for such robots to take on a wide range of everyday tasks, they need to be capable of sophisticated object manipulation. Many vital higher-level tasks will rely on a robot's capability to perform in-hand manipulation by reorienting objects while maintaining the grasp. Out of all the grasping and manipulation tasks, in-hand manipulation is among the ones that require the most dexterity.

A background of in-hand manipulation literature is presented in \cite{yuan2020hand} and a more extensive review of robot hands and graspers is given in \cite{piazza2019century}. 
Common approaches to designing robotic grippers that can perform in-hand manipulation are:
anthropomorphic hands which take advantage of intrinsic human dexterity, but due to the high number of degrees of freedom are complex and expensive \cite{diftler2011robonaut} \cite{company2003design} \cite{grebenstein2011dlr}; 
under-actuated hands which passively conform to objects, achieving good grasp stability, but at the cost of the controllability needed to perform many in-hand manipulation tasks \cite{ma2014underactuated} \cite{dollar2010highly} \cite{Stuart2017ocean} \cite{rojas2016gr2}; 
grippers with active surfaces (such as conveyors) which allow for the object to be manipulated without changing grasp pose, but with a fixed conveyor orientation limiting possible motions \cite{govindan2019multimodal} \cite{tincani2012velvet}  \cite{ma2016hand}. 

\begin{figure}[tb]
      \centering
      \includegraphics[width=3in]{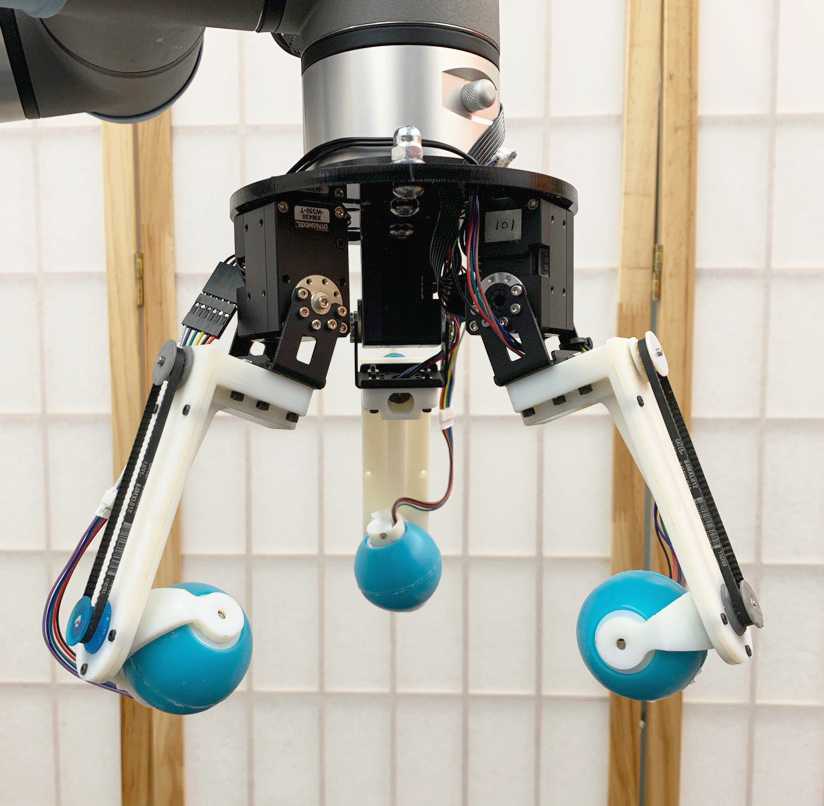}
      \caption{The Roller Grasper V2 prototype mounted on a UR-5 robot arm. Three fingers, each with three degrees of freedom, can grasp and reorient an object using rolling contact with the rubber coated rollers}
      \label{fig:cover_photo}
      \vspace{-5mm}
    \end{figure}

We developed a novel grasper design using articulated, actively driven spherical rollers located at the fingertips, shown in Fig. \ref{fig:cover_photo}. By incorporating continuous rotating mechanisms, it is possible to create graspers that are highly capable but relatively simple by design. The active surface achieved by rolling and reorientation of the spherical rollers allow the grasper to perform in-hand manipulation without the need for finger gaiting. Rolling contact on the object can be viewed as a motion that continuously breaks contact with the object while simultaneously re-establishing contact at adjacent locations. The ability to reorient an object to any pose also lessens the need to use externally actuated degrees of freedom (e.g. actuation of the robotic arm and wrist) which simplifies the control scheme. More importantly, the spherical design of the finger tips allows for stable grasps independent from the roller orientations, eliminating the need to analyze grasping modes for different combinations of roller orientations.

Learning robust policies for in-hand manipulation has been a long-standing challenge in robotics due to the complexity of modelling the object and grasper contacts and the difficulty of controlling finger motion in long and complicated manipulation sequences. Deep Reinforcement Learning (DRL) has been used to learn dexterous manipulation\cite{kumar2016optimal}\cite{andrychowicz2020learning}. However, learning to hold the object
firmly and stably while transforming the object with deep reinforcement learning requires many more training episodes and a carefully designed reward function. To overcome these issues, we used an imitation learning based approach to learn a control policy in order to arbitrarily transform an object while holding it. We demonstrated the effectiveness of this learned policy both in simulation and in real world experiments.

Our in-hand manipulation system consisted of a 3-fingered grasper with spherical rollers at the fingertips, an overhead RGBD camera, and objects with QR-tags on all faces. A handcrafted control policy and an imitation learning policy were developed to perform complex in-hand object transformations. To the best of our knowledge, this work is the first attempt at developing a grasper with active surfaces at the fingertips that transforms grasped objects through an imitation learning policy. The paper is structured as follows: we first discuss our previous iteration of this robotic grasper as well as other design and algorithmic approaches to robotic grasping/in-hand manipulation (Section \ref{sec:relatedwork}). Section \ref{sec:design} then briefly describes the hardware. Section \ref{sec:tech} discusses the formulation of the handcrafted control policy as well as the imitation learning approach. The paper then provides an overview of the experimental setup in simulation and in real life, and concludes by reporting and discussing the results (Section \ref{sec:exp} and Section \ref{sec:result}).

\section{Related Work}
\label{sec:relatedwork}
\subsection{Previous Grasper Design}

Our previous work \cite{yuan2020hand} used articulated, actively driven cylindrical rollers at the fingertips of a grasper 
to explore imparting motion within a grasp using active surfaces. The grasper used three modular 3-DoF  fingers, and demonstrated full 6-DoF spatial manipulation of objects including a sphere, cube, and cylinder, as well as various grasping modalities. One limitation of the previous roller grasper is the grasp stability. Due to the cylindrical design of the finger tips, several grasping configurations are unstable, resulting in undetermined manipulation behaviors. The redundant combinations of grasping configurations also complicates the control scheme as the configuration used is dependent on specific manipulation tasks and the object being manipulated.

    \begin{figure*}[!th]
      \centering
      \includegraphics[width=6.1in]{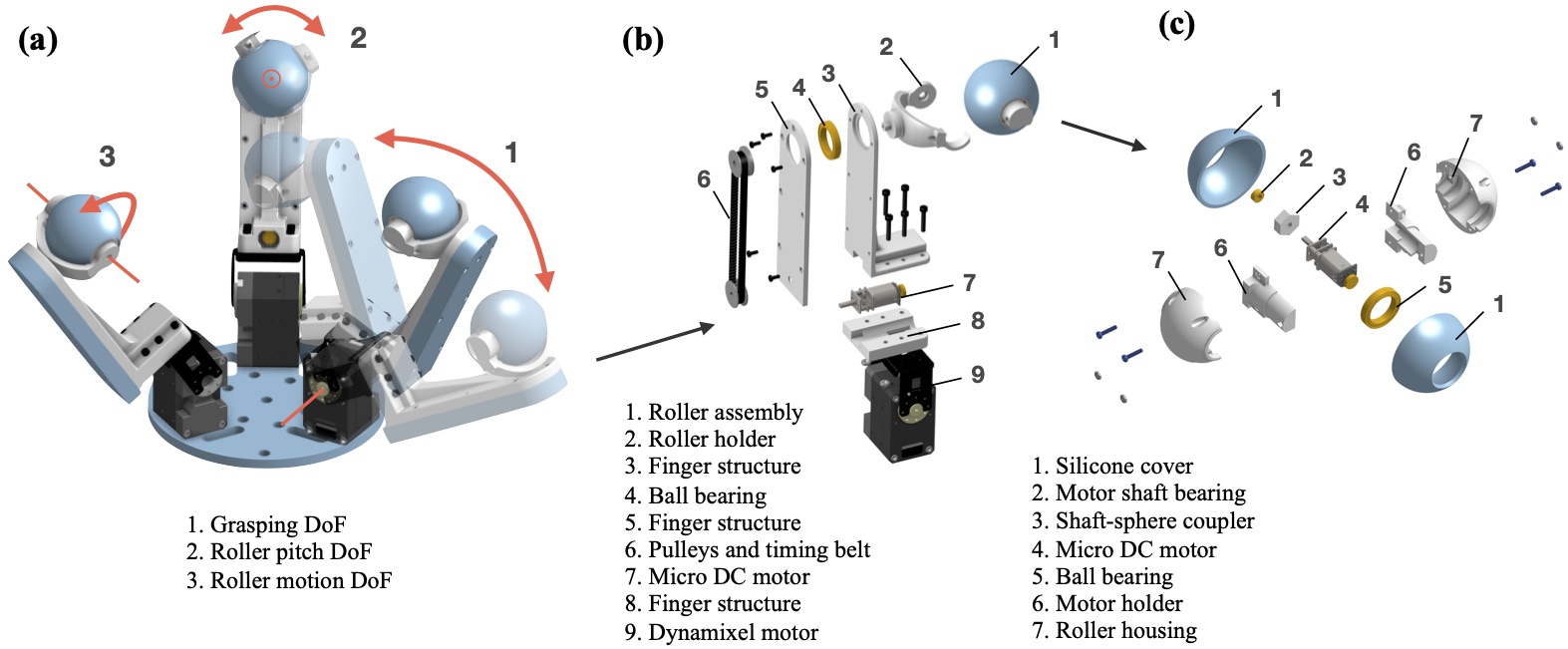}
      \caption{A CAD model of the grasper: (a) the three degrees of freedom each finger has (b) an exploded view of each finger (c) an exploded view of the roller assembly that contacts the object being manipulated}
      \label{fig:full_cad}
      \vspace{-3mm}
    \end{figure*}

\subsection{In-Hand Manipulation}
 In-hand manipulation is an age-old question in robotics with a rich literature, from two-fingered grippers~\cite{chavan2018hand}~\cite{cruciani2018dexterous}, to dexterous hands~\cite{kumar2016optimal}.
We briefly review the relevant in-hand manipulation works in this subsection. To achieve in-hand manipulation, multi-fingered dexterous hands utilize the redundancy of the fingers to move the object while holding it. Under-actuated hands are able to leverage model based control~\cite{liarokapis2017deriving} for in-hand manipulation tasks. There are also task-specific designs of under-actuated hands which enable a limited set of repositioning skills~\cite{7576796}~\cite{bircher2017two}. 

Other approaches to in-hand manipulation have been explored which rely on gravity with controlled slip \cite{karayiannidis2015hand} \cite{brock1988enhancing}, induced accelerations \cite{dafle2014extrinsic} \cite{shi2017dynamic}, or the environment\cite{chavan2018hand} \cite{chavan2015prehensile}  \cite{eppner2015exploitation} to reduce the dexterity required of the hand. However, such approaches require complex control and modeling schemes or dependency on available environmental geometry.

Contrary to modeling the complex dynamics involved in grasping and object reorientation, some researchers have opted to use reinforcement learning (RL) to search for optimal policies. This is especially useful when using underactuated graspers or graspers with high DoF's. In \cite{van2015learning} an underactuated grasper with tactile sensors on the fingertips was used to horizontally slide a wooden cylinder back and forth by rolling it along each finger. The learned policy was evaluated on cylinders of different masses, sizes, and friction coefficients. They found that the policy performed better than a hard-coded control policy, which was used as a baseline, but still struggled with cylinders with low-friction properties. DRL has also been implemented successfully on the 24 DoF Shadow Hand for dynamically moving a cylinder in hand and for arbitrarily reorientating a cube using visual information \cite{kumar2016optimal}\cite{andrychowicz2020learning}. However, our grasper needs to maintain hold of the object solely through friction at the roller contact during the manipulation process. This means that a tiny perturbation of the fingertip could possibly break the contact and lead to a dropped object. Therefore, the space of successful policies is incredibly small relative to the entire policy space. Since exploration is necessary in any DRL problem, it is difficult for the algorithm to converge to the optimal policy. To avoid this problem, we instead adopted an imitation learning method, which will be discussed in the next section.

\subsection{Imitation Learning Methods}
Imitation learning aims to learn control policies by observing expert demonstrations. There are in general two types of approaches to tackle an imitation learning problem. \textit{Behaviour cloning} aims to train the agent to learn a mapping from observations to actions given demonstrations, in a supervised learning fashion~\cite{pomerleau1989alvinn}~\cite{ross2011reduction}. Another approach is \textit{Inverse Reinforcement Learning}~\cite{ng2000algorithms}~\cite{shao2020concept}, which attempts to learn a reward function that describes the given demonstrations. Our method falls under the scope of \textit{Behavior Cloning} which has led to many successes in robotics~\cite{billard2004discovering}~\cite{pastor2009learning}; our approach is based on one \textit{Behaviour Cloning} method called DAgger~\cite{ross2011reduction}. 
To tackle the problem of generating expert demonstrations, we also develop a method to accumulate the demonstration examples iteratively starting from a few expert demonstrations.

\section{Design}
\label{sec:design}
\subsection{Hardware Design}
\begin{figure}[tb]
      \centering
      \includegraphics[width=3.4in]{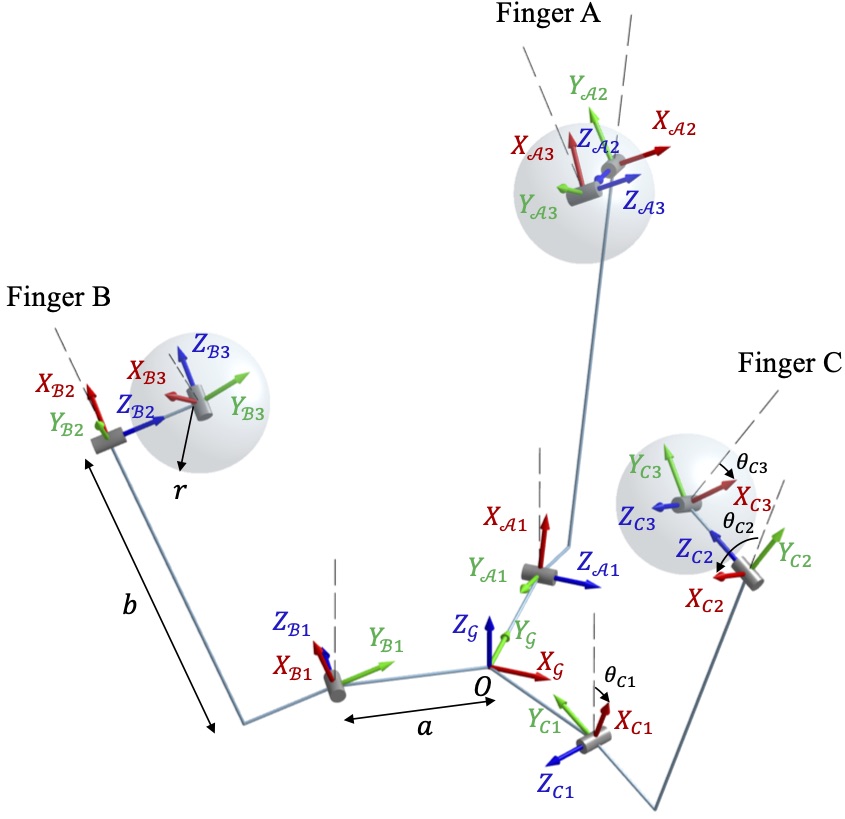}
      \caption{Reference frames of each of the three joints within each finger, relative to the base coordinate system. Dashed lines indicate neutral positions from which joint angles are measured. }
      \label{fig:coord_sys}
      \vspace{-5mm}
    \end{figure}

The gripper (Fig. \ref{fig:full_cad}(a)) consists of three fingers, each having three degrees of freedom (DoF). The first DoF is at the base of each finger and consists of a revolute joint directly driven by a Robotis Dynamixel XM430-W350 smart actuator. The other two DoF are located at each fingertip, and are responsible for steering and rolling. 
The second joint shown in Fig. \ref{fig:full_cad}(a) is orthogonal to the first DoF, and is driven by a micro DC motor with built-in gearbox and quadrature encoder (Servocity No.638099). For a compact form-factor, this actuator is located remotely from the axis of rotation through a timing belt (Fig. \ref{fig:full_cad}(b)), and allows the roller assembly to be pitched up to 180 degrees.
The final DoF is actuated using the same type of geared motor but housed inside the roller assembly (Fig. \ref{fig:full_cad}(c)), allowing it to perform continuous rotation of the spherical contact surface without its cables winding. The roller is encased in a pair of 2mm-thick semi-spherical silicone covers to provide a high-friction surface for grasping and manipulation. The silicone covers are molded from Mold Star 16 Fast (SmoothOn), cast in a 2-part 3D printed mold (Objet Vero white, glossy surface finish), with a coat of Ease Release 2000 release agent. They are adhered to the 3D printed roller surfaces using silicone compatible tape (McMaster 
No.7213A31)
The reference frame of the grasper is shown in Fig. \ref{fig:coord_sys} and the key physical parameters are listed in Table \ref{table_properties}. 

\begin{table}[tb]
    \caption{\sc{Physical Properties}}
    \vspace{-5mm}
    \label{table_properties}
    \begin{center}
    \begin{tabular}{|c |c |}
        \hline
        Property & Value \\
        \hline
        Link $a$ (referred in Fig. \ref{fig:coord_sys})& $48mm$\\
        Link $b$ (referred in Fig. \ref{fig:coord_sys})&  $122mm$\\
        Link $r$ (referred in Fig. \ref{fig:coord_sys})& $21.5mm$\\
        Whole grasper weight & $700 g$\\
        Maximum normal force at the fingertip & $33.6 N$\\
        Maximum roller shear force & $16.4 N$\\
        \hline
    \end{tabular}
    \end{center}
    \vspace{-5mm}
\end{table}

The design allows for unbounded reorientation of the grasped object, while the range of translation is determined by various factors such as the physical dimensions, shape, in-hand orientation, mass, and resulting friction coefficient of the grasped objects. 

\subsection{System Architecture}
\begin{figure}[tb]
      \centering
      \includegraphics[width=3.2in]{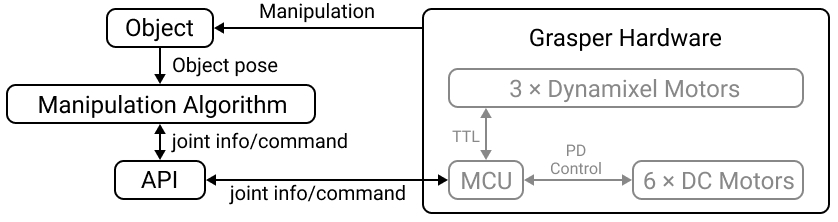}
      \caption{System architecture}
      \label{fig:system_diagram}
      \vspace{-5mm}
    \end{figure}

The system architecture used to operate the gripper is shown in Fig. \ref{fig:system_diagram}. A high-level API was developed to interface between the low-level (hardware) information and the manipulation algorithm. Information transferred during the bidirectional communication includes positions for each joint of the fingers, the current limit of the base joint, as well as the control parameters for controlling the motors. Current to the Dynamixel motors is used to set the stiffness of the base joints and 
estimate the force exerted by the object on each finger during manipulation.

A Teensy 3.6 microcontroller is used to handle communication with the high-level API as well as low-level control of the motors. The six micro gearmotors located at the intermediate joints and within the rollers are controlled by PD position controllers handled by the microcontroller. The Dynamixel motors each run local PD control and communicate with the Teensy microcontroller through a TTL half-duplex asynchronous serial communication protocol. 

\section{Technical Approach}
\label{sec:tech}

In this section we begin by describing the analytical approach that led to development and implementation of a handcrafted control policy. We then discuss how this control policy was fine-tuned based on simulation results in order to act as an expert trajectory generator for the imitation learning. The final subsection describes the imitation learning formulation.

Initially, several DRL methods were used in place of imitation learning. From these preliminary experiments we learned that the space of successful policies, not necessarily optimal, was especially small. This becomes a critical problem in DRL given large policy spaces since significant exploration is necessary. Instead, we elected to pursue imitation learning which proved to be both effective and efficient.

\subsection{Analysis}

Manipulating an object through rolling contact can be viewed as navigating the rollers on the object. Therefore, an object transformation can be achieved by navigating the rollers from their initial contact locations to the desired final contact locations. While there is no unique solution for the paths the rollers take during their navigation for given initial and final grasping poses, it is possible to solve, on this non-holonomic system, for the instantaneous joint velocities based on known object geometry, object pose, object velocity, and contact conditions (rolling without slipping).

Object geometry and pose are necessary to calculate the contact locations, which determine the relationship between joint motions and given contact motions on the rollers (the contact jacobian matrix). This information can subsequently be used to map the desired object motion to the motions at the contact point on the object. Applying the contact condition of rolling without slipping means that the contact velocity on the object and the contact velocity on the roller are equal. Thus, inverse kinematics can be used to calculate the joint velocities required for a desired object motion, a process similar to the calculation of geometric stiffness for grasping presented in \cite{cutkosky1989computing}. 

    \begin{figure}[b]
      \centering
      \includegraphics[width=1.7in]{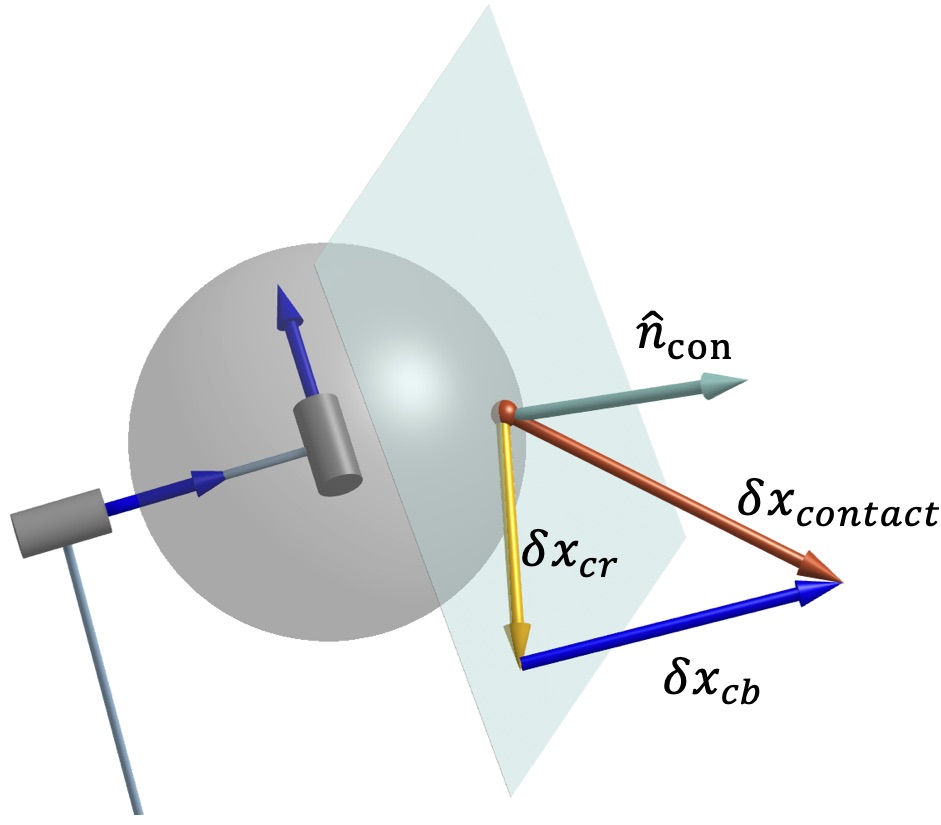}
      \caption{Contact motion breaks down to two components}
      \label{fig:v_con_split}
      \vspace{-5mm}
    \end{figure}

The problem is formulated as follows: given the object's initial and target position and orientation, compute the contact location and contact motion on the roller, and then compute the pivot joint orientation and motions of the base and roller joints.

Given the desired motion of the object and known contact locations, we can obtain the contact motion by

    \begin{equation} \label{equ: contact_jacobian_obj}
         \delta x_{contact} = J_{obj} \delta {x}_{obj}
    \end{equation}

\noindent where $J_{obj}$ is the jacobian matrix mapping object motion to the motion at the contact frame. 

On the other hand, with the contact motion $\delta x_{contact}$, the object position $x_{obj}$, and roller position $x_{roller}$ all known, the contact motion can be interpreted as the motion at the contact location due to the movement of the 3 joints:
    \begin{equation} \label{equ: contact_jacobian}
         \delta x_{contact} = J_{\theta} \delta \theta
    \end{equation}
Where $J_{\theta}$ is the contact jacobian matrix for mapping finger joint motions to motions in the contact coordinates, and $\delta \theta$ is a vector of joint motions of the finger. In many robotics applications $\delta \theta$ is determined directly by solving (\ref{equ: contact_jacobian}). However, this method is not particularly applicable to this gripper design for the following reasons:
\begin{enumerate}
    \item The contact locations are very close to the pivot joint so the finger is always close to a singularity. 
    \item The pivot joint has a joint limit of $[-\frac{\pi}{2},\frac{\pi}{2}]$ meaning that in many cases the instantaneous velocity required from the pivot joint to achieve the desired contact velocity cannot be reached.
\end{enumerate}

Therefore, instead of carrying out the full inverse kinematics, we divide the contact motion into two components: the component resulting from the motion of the base joint, $\delta x_{cb}$, and the component resulting from the rolling, $\delta x_{cr}$. The pivot joint is used to reorient the roller so that the rolling direction at the contact point is aligned with $\delta x_{cr}$. This approximation is sufficient because (1) when the contact locations are close to the pivot axis, the pivot motion does not have a significant impact on the object motion, and (2) the object being grasped is over constrained by the three fingers, so using the soft finger model \cite{mason1985robot} to approximate roller contacts, the torque exerted by pivoting is compensated by the lateral frictions from the two other fingers. Thus, the singularity is advantageous in that it enables sole use of the pivot joint to align the roller in order to achieve $\delta x_{cr}$.

We developed a handcrafted control strategy based on the above formulation with some modifications that fit the gripper design. The handcrafted control policy tested in simulation and on the hardware assumes that the object geometry is spherical with radius $R$. Specifically, the below calculations are for a single finger, but can be applied to the two other fingers as well. For the simplicity of notation, we use subscript $1$, $2$, and $3$ to represent the base joint, pivot joint, and roller joint, respectively, and leave out the subscript $\mathcal{A}$, $\mathcal{B}$ or $\mathcal{C}$ because the following derivation would be identical for each finger. 

In order to obtain $\delta x_{cb}$ and $\delta x_{cr}$, we need to project $\delta x_{contact}$ onto ${Z}_2$ and the contact plane (Fig. \ref{fig:v_con_split}). The contact plane can be described by its normal vector $\vec{n}_{con}$:
    \begin{align}
         \hat{n}_{con} &= \frac{x_{obj} - x_{roller}}{||x_{obj} - x_{roller}||_2}
    \end{align}

\noindent Using the fact that $\delta x_{cr}$ is orthogonal to $\hat{n}_{con}$ and is also in the plane formed by $\delta x_{cb}$ and $\delta x_{contact}$, we can compute the direction of $\delta x_{cr}$ as:
    \begin{equation}
         \widehat{\delta x}_{cr} = \frac{(\delta x_{cb}\times \delta x_{contact}) \times \hat{n}_{con}}{||(\delta x_{cb}\times \delta x_{contact}) \times \hat{n}_{con}||_2}
    \end{equation}
It is also easy to find that the direction of $\delta x_{cb}$ aligns with $Z_2$:
    \begin{equation}
         \widehat{\delta x}_{cb} = Z_2
    \end{equation}
Projecting $\delta x_{contact}$ onto $\widehat{\delta x}_{cb}$ and $\widehat{\delta x}_{cr}$ gives $\delta x_{cb}$ and $\delta x_{cr}$. It is equivalent to solving $\alpha$ and $\beta$ in equation: 

    \begin{equation}    \label{equ:v_con_split}
        \delta x_{contact} = \alpha \widehat{\delta x}_{cb} + \beta \widehat{\delta x}_{cr}
    \end{equation}
 By cross multiplying $\widehat{\delta x}_{cb}$ and $\widehat{\delta x}_{cr}$ to (\ref{equ:v_con_split}), respectively, we can solve for $\alpha$ and $\beta$, and the resulting $\delta x_{cb}$ and $\delta x_{cr}$ are shown below. Note that $\hat{n}_z$ is only used to extract the magnitude from the cross products. 

    \begin{align}
        \hat{n}_z &= \frac{\widehat{\delta x}_{cr} \times \widehat{\delta x}_{cb}}{||\widehat{\delta x}_{cr} \times \widehat{\delta x}_{cb}||_2}   \\
        \delta x_{cb} &= \frac{\hat{n}_z \cdot (\widehat{\delta x}_{cr} \times \delta x_{contact})}{\hat{n}_z \cdot (\widehat{\delta x}_{cr} \times \widehat{\delta x}_{cb})} \widehat{\delta x}_{cb}    \\
        \delta x_{cr} &= \frac{\hat{n}_z (\cdot \widehat{\delta x}_{cb} \times \delta x_{contact})}{\hat{n}_z \cdot (\widehat{\delta x}_{cb} \times \widehat{\delta x}_{cr})} \widehat{\delta x}_{cr}
    \end{align} 

\noindent The joint velocity of base joint ($\omega_1$) and roller joint ($\omega_3$) can be calculated using inverse kinematics.

The final step is to calculate the pivot angle $\hat{\theta}_2$ to align the rolling direction with $\widehat{\delta x}_{cr}$:
    \begin{align}
        Z_3 = \pm \frac{{Z}_2 \times \widehat{\delta x}_{cr}}{||{Z}_2 \times \widehat{\delta x}_{cr}||_2}
    \end{align}
Note that because there is a joint limit at the pivot, we are limiting $Z_3$ to always have a component along the ${Z}_0$ direction. The pivot angle $\theta_2$ can be calculated by 
    \begin{equation}
        \theta_2 = \arccos\left(\frac{Z_1 \cdot Z_3}{||Z_1||_2||Z_3||_2}\right)
    \end{equation}

The above process only describes how each joint should move given the instantaneous desired velocity of the object; no path planning is used for navigating the rollers given the initial and target poses of the object. In our case, we simply compute the difference between the initial and target pose, and set the desired velocity equal to a scaling factor multiplied by this computed difference: $\delta{x}_{obj} = \lambda \Delta x_{obj}$. This method works very well for convex objects whose radii of curvature do not change drastically. Experimental validation of manipulating a cube is shown in Section \ref{sec:exp}.

\subsection{Handcrafted Control Policy}

The handcrafted control policy is formulated according to the results from the previous section. Given the current object position and orientation, the target object position and orientation, and the current joint values, the positions of all nine joints are calculated for the following time step. One difference between the implemented policy and the theoretical policy is that the base joint velocities are not controlled based on the derivation above. Instead, they are position controlled to a setpoint in order for the rollers to maintain contact with the object. This is because we are focusing on the rotation of the object instead of translation. The main purpose of the base joint in our case is to keep the rollers in contact with the object. The analytical approach presented in the previous section performs both translation and rotation. The translation capability of the grasper is demonstrated in the complementary video through scripted movements. However, the translation capability is relatively limited compared to the rotation capability, which is why we decided to focus on object rotation in the control policy. 

\subsection{Imitation Learning Algorithm}
We adopted imitation learning, specifically \textit{Behavior Cloning} in order to learn how to transform an object.  
The optimal policy was learned through expert demonstrations characterized by the above handcrafted control policy.
\textit{Behavior Cloning} is particularly useful when it is difficult to explicitly specify the reward function. 

In our case, the state space is defined as $|\mathcal{S}| \in \mathbb{R}^{35}$, and consists of the following: the current state of the grasper ($s_1 \rightarrow s_9$), current object position ($s_{10} \rightarrow s_{12}$), current object quaternion ($s_{13} \rightarrow s_{16}$), previous object position ($s_{17} \rightarrow s_{19}$), previous object quaternion ($s_{20} \rightarrow s_{23}$), object termination position ($s_{24} \rightarrow s_{26}$), object termination orientation in angle-axis representation ($s_{27} \rightarrow s_{29}$), object initial position ($s_{30} \rightarrow s_{32}$), and object initial orientation in angle-axis representation ($s_{33} \rightarrow s_{35}$). The action space is defined as $|\mathcal{A}| \in \mathbb{R}^{9}$ and contains the nine joint positions ($a_1 \rightarrow a_9$). 

We constructed a deep neural network to determine the actions for each of the nine gripper joints. The network consisted of three fully connected hidden layers with leaky ReLU activations, except at the output layer, and 256 nodes in each hidden layer.

The handcrafted control policy from the previous section was used to generate $N$ expert trajectories, which are simply a series of state-action pairs. The $i$th trajectory is defined as: 

\begin{equation}
    T^{(i)} = [(s_0^{(i)}, a_0^{(i)}), (s_1^{(i)}, a_1^{(i)}),\dots]
\end{equation}

We first trained our policy $\pi^0(s^i)$ to predict the expert action by minimizing the loss $L$ between $\pi^0(s^i_j)$ and $a^i_j$ in a supervised approach:

\begin{equation}
    L_j^i = \|\pi^0(s^i_j) - a_j^i\|_2, \:\forall\: s_j\in |\mathcal{T}^i|,\: i\in N
\end{equation}

\noindent Subsequent policy updates are computed according to DAgger \cite{ross2011reduction}.

We also implemented a method to increase the number of expert demonstrations iteratively. For a given object, every object transformation trajectory is specified by a 12-dimensional vector containing the object starting and target position and orientation: $(x^i_{obj,s}, q^i_{obj,s}, x^i_{obj,t}, q^i_{obj,t})$. By imitating the expert demonstration examples, we learn a policy supported by these expert demonstrations. This control embedding is able to interpolate between known trajectories using a nearest neighbor policy in order to generate trajectories for previously unseen transformations.

Based on the learned policy, we  accumulate nearby transformations. Let $\mathcal{D}=(x^i_{obj,s}, q^i_{obj,s}, x^i_{obj,t}, q^i_{obj,t})$ represent the transformations which our policy already knows. If an interpolated trajectory transformation is nearby to one of the transformations in $\mathcal{D}$, we add the transformation to $\mathcal{D}$, and then continue to train our policy based on the transformation demonstrations in $\mathcal{D}$. Through this fashion, we kept growing and maintaining a traversable graph of transforming the object between various poses, similar to~\cite{cruciani2019dual}. The learned control policy could also be treated as an efficient initialization for a deep reinforcement learning approach or motion planning for in-hand manipulations.

\section{Experiments}
\label{sec:exp}

\subsection{Simulation Experiments}

\begin{figure}[b]
  \centering 
  \includegraphics[width=1.7in]{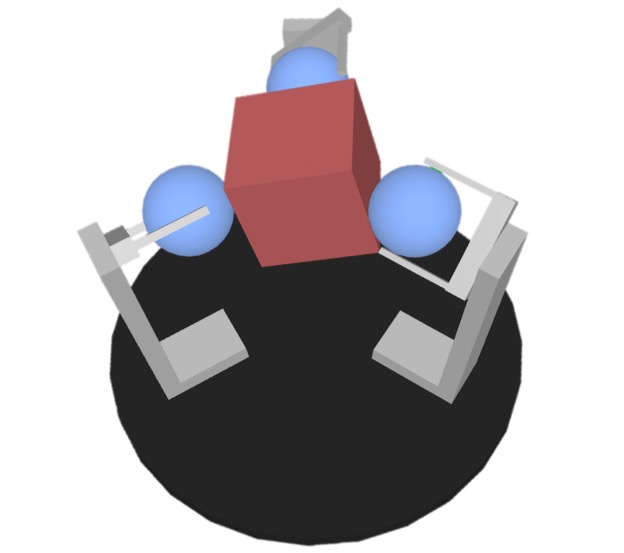} 
  \caption{A visualization of our simulation in Mujoco. In  simulation the gripper maintains the same physical design as the real gripper.}
  \label{fig:sim}
  \vspace{-3mm}
\end{figure}

Mujoco 2.0 was used to simulate both the handcrafted control policy and the learned policy before transferring them to the physical setup (see Fig. \ref{fig:sim}). By default, Mujoco assumes a soft-contact model which leads to relatively slippery contacts, but the solver parameters (\textit{solref} and \textit{solimp}) were changed to have a harder contact model in order to better model the friction between the object and the spherical rollers. An elliptical friction cone was used along with the Newton solver to evaluate these constraints.

The gripper was modeled in the software, and the following setup was specified. The base joints were position controlled, and had their force output clamped; these settings allowed the fingers to act compliantly and conform to an object's shape as it was reoriented, and stabilized the grasp. The pivots had a rotation range from $[-\frac{\pi}{2}, \frac{\pi}{2}]$ (with the zero position corresponding with the roller axis aligned along the length of the finger) in order to represent the limited range in the physical setup due to motor wires. A pivot rotation threshold of $3^\circ$ per time step was used to prohibit quick jumps between the two rotation limits in favor of smoother motion that stabilized the simulation. 

    \begin{figure}[tb]
      \centering
      \includegraphics[width=2.5in]{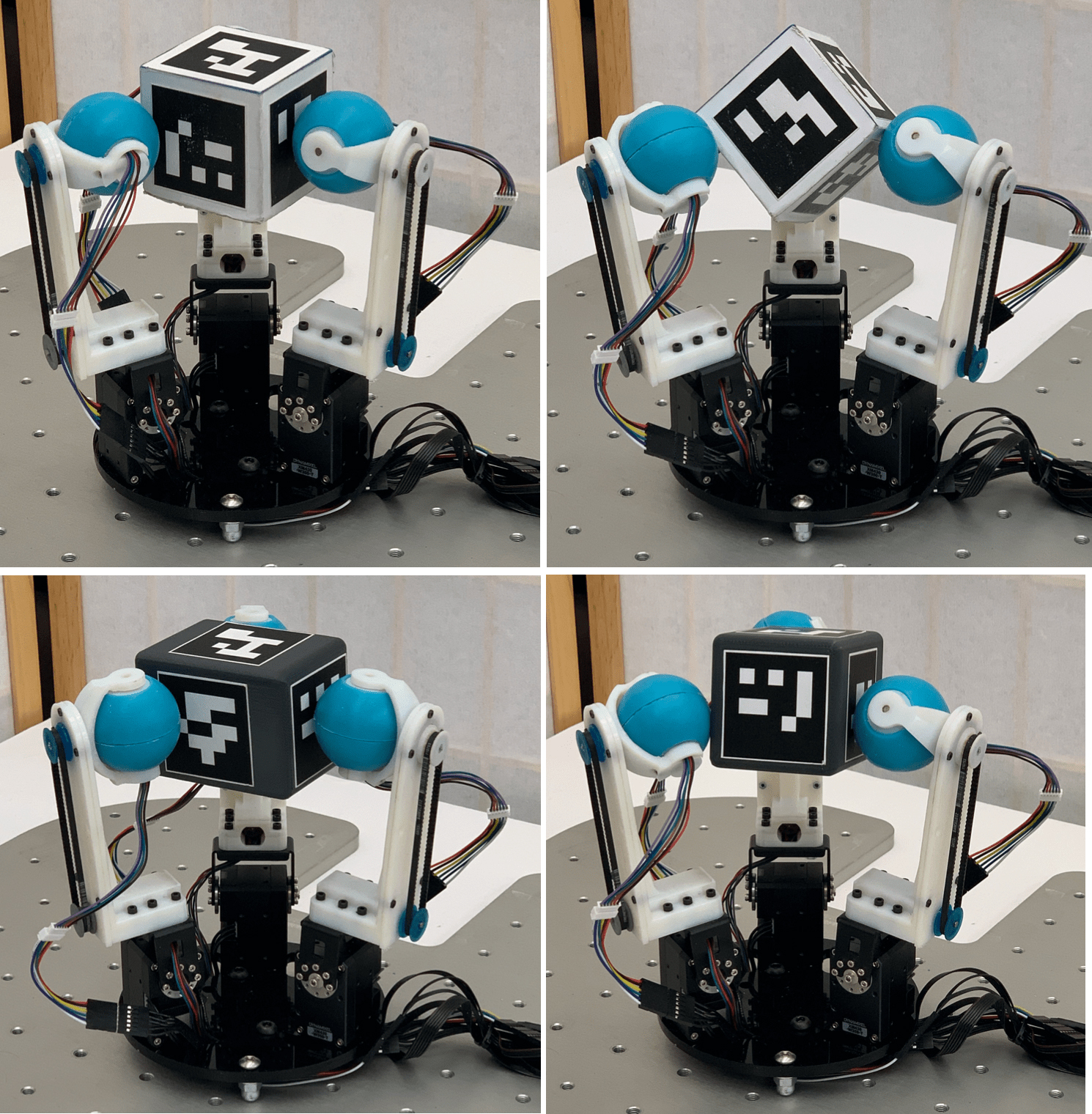}
      \caption{Top row: the cube being manipulated from a starting position shown on the left. Bottom row: the novel object, a rectangular prism being similarly manipulated}
      \label{fig:exp_photo}
      \vspace{-5mm}
    \end{figure}

Sensors were placed at all nine joints to read joint positions and two sensors were placed on the object to obtain its orientation (quaternion) and 3D position relative to the global frame. The learned policy outputted values for all nine joints. The handcrafted control policy only used the latter two sensors and the base joint positions to calculate the output velocities for the pivots and rollers at every time step.

Experiments were run by specifying a desired orientation in angle-axis representation $\left([x, y, z]^T,\:\theta\right)$. Most experiments were carried out with a 6 cm cube with a mass of 32.4 grams that had a starting quaternion of $q_0 = [1, 0, 0, 0]$. Object rotation angles were specified as at least $90^\circ$ in the majority of experiments to ensure that the rollers had to traverse across the edges of the cube. Rotation axes ranged from more vertical axes -- which were generally easier for the simulation to realize -- to purely horizontal axes which were more difficult.

\subsection{Experimental Setup}

The experimental setup included the grasper, an overhead Intel Realsense D400series RGBD camera, and various 3D printed objects including a cube, a cube with filleted edges, and spheres of various masses and sizes. The handcrafted control policy was able to be run both open-loop and closed-loop. Since object orientation and position were used as input to the control policy, only the cube was run in the closed-loop operation with QR-tags on all 6 faces. Open-loop runs with the  spheres were used to qualitatively verify the handcrafted control policy on the hardware.

At the start of each experiment the fingers would spread out, allowing a human operator to hold the cube at approximately the initial position specified for that particular trial. The fingers would then move inwards and grasp the object. At every time step the joint positions, object orientation, and position would be read-in, from which the corresponding joint output would be calculated and sent to the actuators.

\subsection{Evaluation Metric}\label{sec:err}

We adopted the orientation error metric suggested in~\cite{huynh2009metrics}  which can be computed by using the sum of quaternions and normalizing. Given the desired object orientation, $q_{obj,d}$, and the current object orientation, $q_{obj,c}$, the error is calculated as the following:

\vspace{-4 mm}
\begin{multline}
    e_{\omega} =
    100  \frac{\min(||q_{obj,d} - q_{obj,c}||_2,||q_{obj,d} + q_{obj,c}||_2)}{\sqrt{2}}
    \label{eqn_error}
\end{multline}

\section{Results and Analysis}
\label{sec:result}




    \begin{figure}[tb]
      \centering
      \includegraphics[width=3.5in]{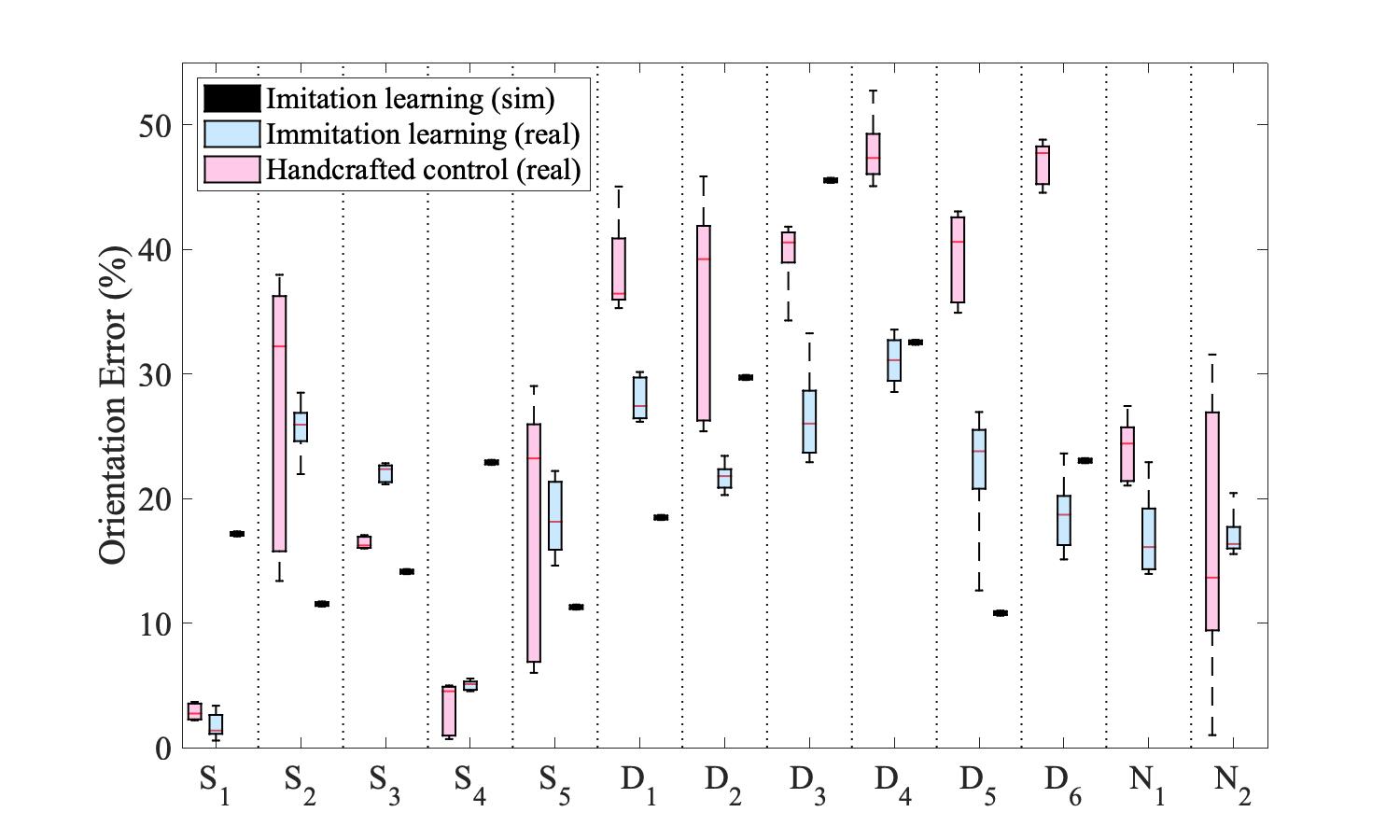}
      \caption{Orientation error for real world experiment results. The vertical axes represents the orientation error defined in~\ref{sec:err}. The horizontal lines represents various experiments. $S$ $D$ and $N$ denote simple manipulations tasks, difficult manipulation tasks and novel object transformations,respectively. Notes: (1) The handcrafted control is designed to stop at a pre-defined error threshold, therefore its orientation error is not meaningful to shown. (2) Imitation learning policy in simulation is deterministic. (3) The novel object transformation was only tested in real-world experiments, thus no simulation results were available.}
      \label{fig:exp_result}
      \vspace{-5mm}
    \end{figure}

The results of the experiments carried out on the hardware are presented in Fig. \ref{fig:exp_result}. All reported percent average errors were calculated according to (\ref{eqn_error}). The experiments were divided up as follows: simple transformations, which had axes of rotations with strong $Z_{\mathcal{G}}$ components, difficult transformations, which had axes of rotations with strong $x$ and $y$ components, and novel object transformations, which consisted of transforming an elongated rectangular prism with filleted edges (6 cm $\times$ 6 cm $\times$ 8 cm) (Fig. \ref{fig:exp_photo}). The reason transformations with strong $Z_{\mathcal{G}}$ components for the rotation axis were easier was due to the three-finger design of the grasper, leading to a much better force closure around $X_{\mathcal{G}}-Y_{\mathcal{G}}$ direction than the $Z_{\mathcal{G}}$ direction. All experiments specified the target as an axis and a $90^\circ$ rotation about that axis. This choice of a rotation angle is described in \textit{Simulation Experiments}. 

\subsection{Simple Cases \& Sim-to-Real Transferring}
The imitation learning method performed better for 3 out of the 5 simple transformation cases ($S_1 \rightarrow S_5$). The average orientation error across these 5 trials shows that the imitation learning method ($15.3\% \pm 2.2\%$) and the handcrafted control policy ($13.6\% \pm 7.2\%$) are comparable when performing object rotations about more vertical axes. However, the handcrafted control policy has more than 3 times the standard deviation of the imitation learning. 

As with any physical system there is a performance difference when compared to the simulation setup. Fortunately, this difference was small with the imitation learning's results in simulation reporting an average error of $15.4\%$. We believe this is due to inclusion of sensor noise in the simulation, as well as fine tuning of the contact model to more closely align with observations of the physical system performance. 

\subsection{Untrained Target Pose Test Cases}
For the difficult transformations ($D_1 \rightarrow D_6$), the imitation learning outperformed the handcrafted policy across all 6 trials by an average percent average error of $16.0\%$. Three of these trials, $D_1 \rightarrow D_3$, including target poses not trained in simulation. Overall, the imitation learning achieved a percent average error of $25.3\% \pm 4.2\%$, while the handcrafted control policy achieved $41.3\% \pm 5.8\%$. Again, the imitation learning was able to demonstrate more stable trajectories despite the sensor noise. 

\subsection{Novel Object Test Cases}
Tests with the novel object ($N_1 \rightarrow N_2$), a rectangular prism, demonstrated similar results to the simple transformation case. The percent average errors were comparable between imitation learning ($17.1\% \pm 3.9\%$) and the handcrafted control policy ($20.3\% \pm 10.3\%$). However, the handcrafted control policy had a much higher standard deviation.


\subsection{Discussion}
A potential explanation for the poorer performance of the handcrafted control policy, especially for the difficult transformations, was that it always generated linear trajectories from the current position to the target. However, in many cases, especially for rotation axes with dominating horizontal components, taking a linear path can lead to the gripper dropping the object. As the object is rotated closer to the desired axis of rotation, one or more of the rollers can roll onto a surface where the contact normal has a $-Z_{\mathcal{G}}$ component; the contact normal can overcome the frictional support provided by another roller and result in dropping the object. This problem can be mitigated by increasing the positional error compensation in the closed-loop control method in order to more actively control the object height. Unfortunately, placing a greater emphasis on controlling the object position was seen to cause previously successful trajectories to fail. No combination of gains was seen to work across all observed successes. On the other hand, the imitation learning was better for a couple of potential reasons: (1) the policy was learned from noisy sensor data which increased its robustness, and (2) the discovery of safe and repeatable trajectories.

\section{Conclusion and Future Work}
\label{sec:conclusion}
This paper presents Roller Grasper V2, a new design for a grasper based on steerable spherical rollers located at the finger tips. The hardware design choices and engineering details were provided. A handcrafted control policy was constructed that utilized the active surfaces of the rollers to transform an object to an arbitrary target pose. This control policy was used to generate expert trajectories in order to develop an imitation learning based policy. This learned policy was able to interpolate between learned trajectories in order to successfully reach new targets; these new trajectories were placed in a buffer of known trajectories in order to develop a traversable graph. Experiments for the handcrafted policy and the learned policy demonstrated the utility of this graph for generating safe trajectories, as the learned policy outperformed the handcrafted policy for all difficult target poses. For simple cases, the two policies performed comparably, except the learned policy had much lower variance in the trajectory. Future work includes reducing the size of the rollers to allow for manipulation of smaller objects, developing a novel mechanism that solves the non-holonomic constraints and incorporating tactile sensing on the rollers to provide high-fidelity feedback.

{\tiny
\bibliographystyle{ieeetr}
\bibliography{root}
}

\end{document}